\documentclass{bmvc2k}

\usepackage{multirow}
\usepackage{amsmath}
\usepackage{amssymb}
\usepackage{amsfonts}
\usepackage{graphicx}
\usepackage{enumitem}
\usepackage{verbatim} 

\usepackage{float}

\newcommand{\networkName}{\textit{Query-based transformer decoder }}
\newcommand{\problemName}{\textit{Temporal Skipping} }
\newcommand{\backbone}{\textit{Feature Extractor} }
\newcommand{\temporaldecoder}{\textit{Temporal Decoder} }
\newcommand{\regressionhead}{\textit{Weight-Score Regression Head} }

\title{Interpretable Long-term Action Quality Assessment}

\addauthor{Xu Dong}{xd00101@surrey.ac.uk}{1}
\addauthor{Xinran Liu}{xl01315@surrey.ac.uk}{1}
\addauthor{Wanqing Li}{wanqing@uow.edu.au}{2}
\addauthor{Anthony Adeyemi-Ejeye}{femi.ae@surrey.ac.uk}{1}
\addauthor{Andrew Gilbert}{a.gilbert@surrey.ac.uk}{1}

\addinstitution{
 University of Surrey \\
 Guildford, UK
}
\addinstitution{
 Advanced Multimedia Research Lab \\
 University of Wollongong \\
 Wollongong, Australia
}

\runninghead{Dong, ET AL.}{Interpretable Long-term AQA}

\begin{document}

\maketitle

\begin{abstract}

Long-term Action Quality Assessment (AQA) evaluates the execution of activities in videos. However, the length presents challenges in fine-grained interpretability, with current AQA methods typically producing a single score by averaging clip features, lacking detailed semantic meanings of individual clips. Long-term videos pose additional difficulty due to the complexity and diversity of actions, exacerbating interpretability challenges. While query-based transformer networks offer promising long-term modelling capabilities, their interpretability in AQA remains unsatisfactory due to a phenomenon we term \problemName, where the model skips self-attention layers to prevent output degradation. To address this, we propose an attention loss function and a query initialization method to enhance performance and interpretability. Additionally, we introduce a weight-score regression module designed to approximate the scoring patterns observed in human judgments and replace conventional single-score regression, improving the rationality of interpretability. Our approach achieves state-of-the-art results on three real-world, long-term AQA benchmarks. Our code is available at: \href{https://github.com/dx199771/Interpretability-AQA}{https://github.com/dx199771/Interpretability-AQA}

\end{abstract}

\section{Introduction}

\label{sec:intro}
\begin{figure}[h]
  \centering
  \includegraphics[width=0.98\textwidth,height=0.43\textwidth,trim=0cm 5.5cm 0cm 0cm]{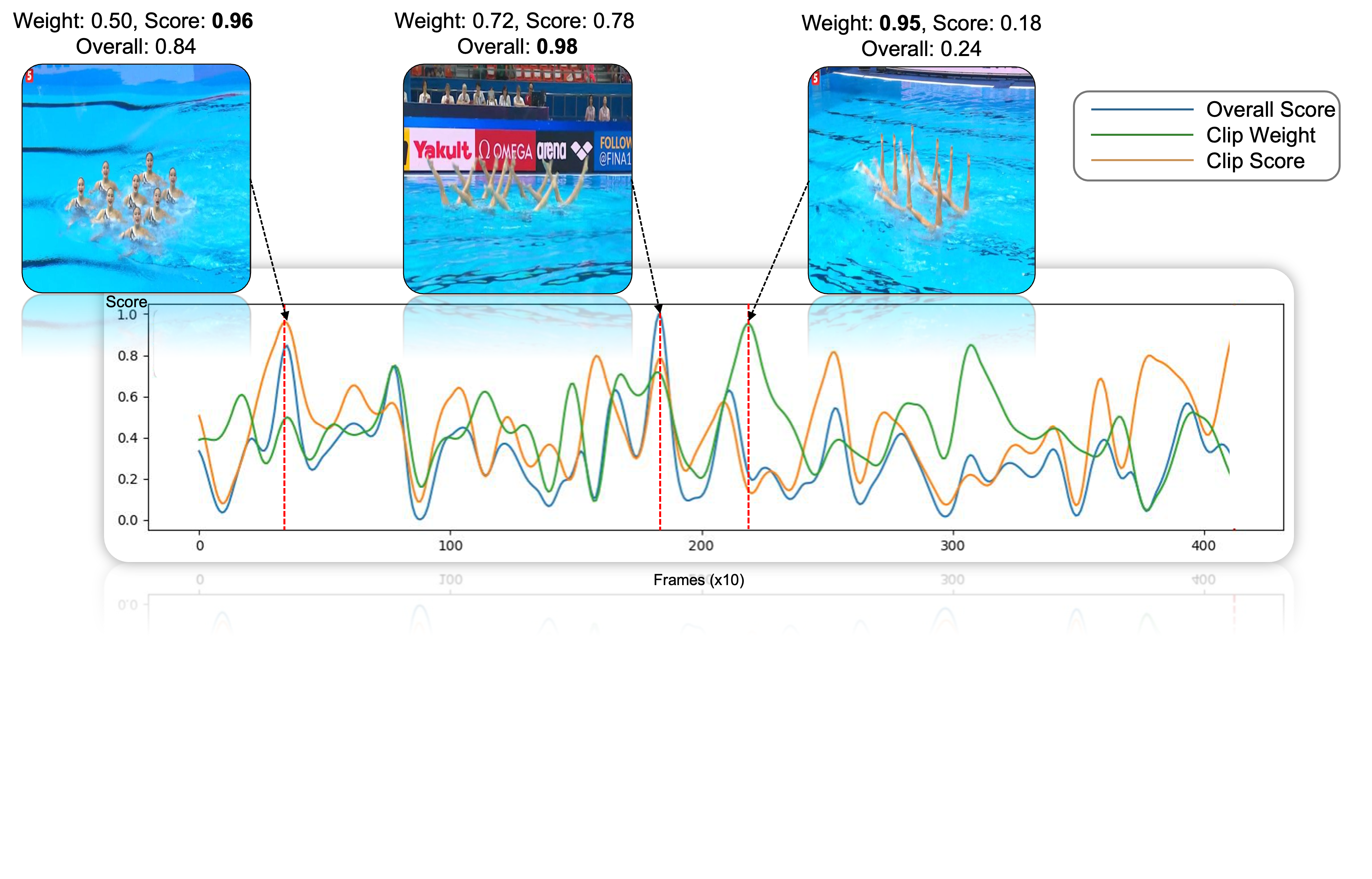}
  \caption{The visualization of the clip-level weight-score regression method illustrates that our network can adhere to the same evaluative logic as human judges in real-world scenarios. The green curve representing weight delineates the significance of the respective action clip, whereas the orange curve for score quantifies the execution quality of the action, the overall score is shown by the blue curve. All scores are normalized to a range of 0 to 1 for easier comparison.}
  
  \label{fig:inter}
  \vspace{-6pt}
\end{figure}
Action quality assessment (AQA) is a task to evaluate how well a particular action is performed. Recently, this problem has garnered increasing interest within the computer vision research community and has a wide range of applications across different real-world scenarios. It is widely used in sports video analysis, including synchronized swimming, figure skating, and gymnastics \cite{parmar2017learning,parmar2019performed,pirsiavash2014assessing,venkataraman2015dynamical,xu2019learning,zeng2020hybrid,parmar2019action}. Providing assistive analysis of athletes' performances and offers objective and precise scoring. Furthermore, AQA can also be used in healthcare \cite{funke2019video,wang2020towards,gao2014jhu}, technical skill training and education purposes \cite{Doughty_2018_CVPR,Doughty_2019_CVPR,li2019manipulation}. 

AQA is typically a score regression task, \cite{parmar2019action,pan2019action,xu2019learning,zhang2024auto,wang2021tsa} directly predicting the final scores for entire action video by aggregating clip features through simple averaging and an MLP regression head. However, a single score fails to provide detailed feedback on the individual components or subtle differences, lacking fine-grained interpretation of temporal sequences. In sports contexts where technical skills and proficiency are crucial, such as gymnastics and artistic swimming, judges typically calculate the final score by weighting each action based on its execution quality and the difficulty factor. Moreover, Prior research on the interpretability of AQA \cite{roditakis2021towards,bai2022action} focused on short-term actions such as diving. These action videos typically last only a few seconds and follow a sequential pattern. Compared to short-term AQA (5 to 10 seconds), long-term AQA (over 120 seconds) tasks are more challenging due to the complexity and diversity of the information and actions involved.



\begin{figure}[t]
    \centering
    
    \subfigure[]{
        \includegraphics[width=0.23\linewidth,trim=0cm 0cm 0cm 0cm]{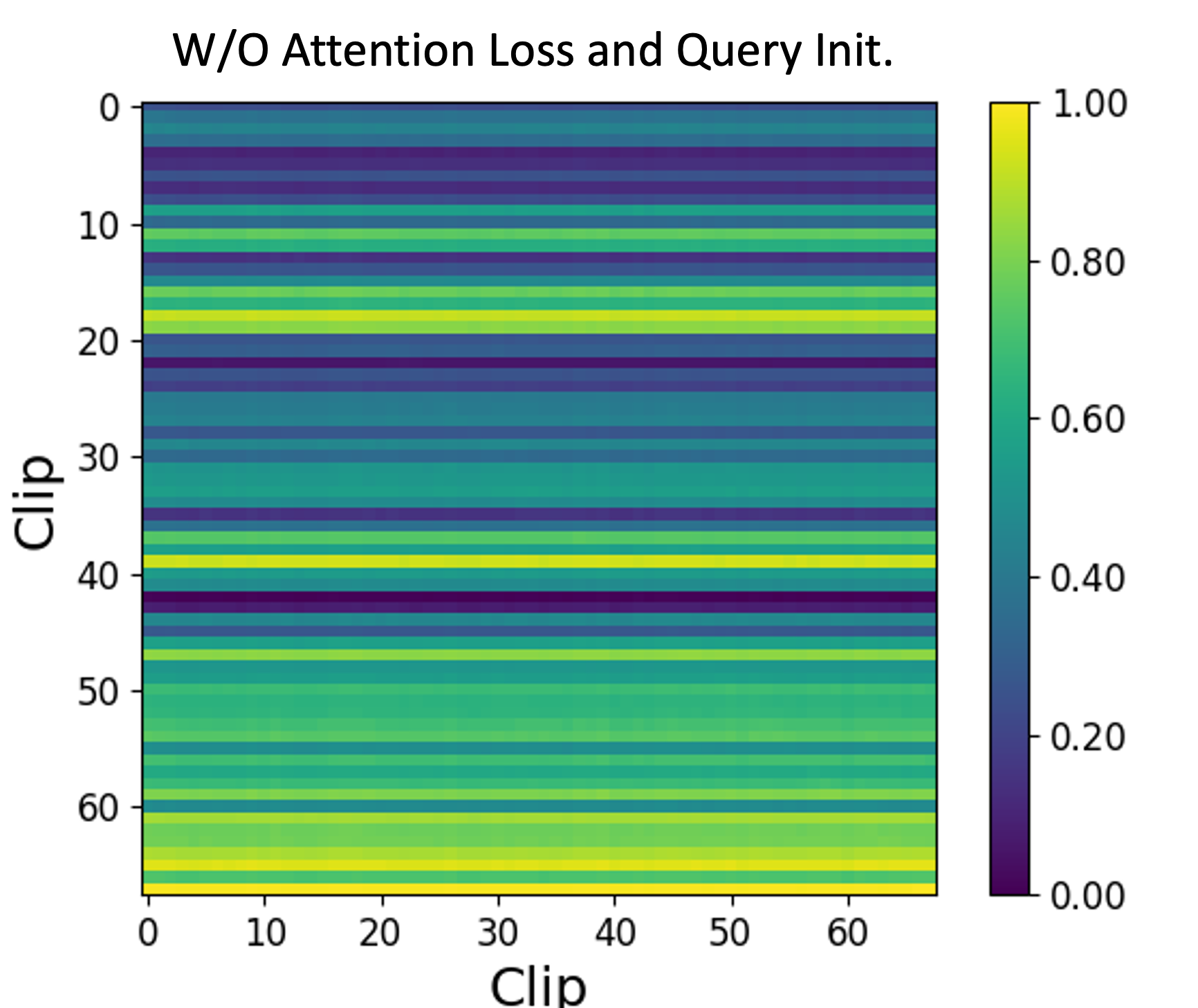}\
        \label{fig:3}
    }
    \subfigure[]{
        \includegraphics[width=0.23\linewidth,trim=1cm 0cm 0cm 0cm]{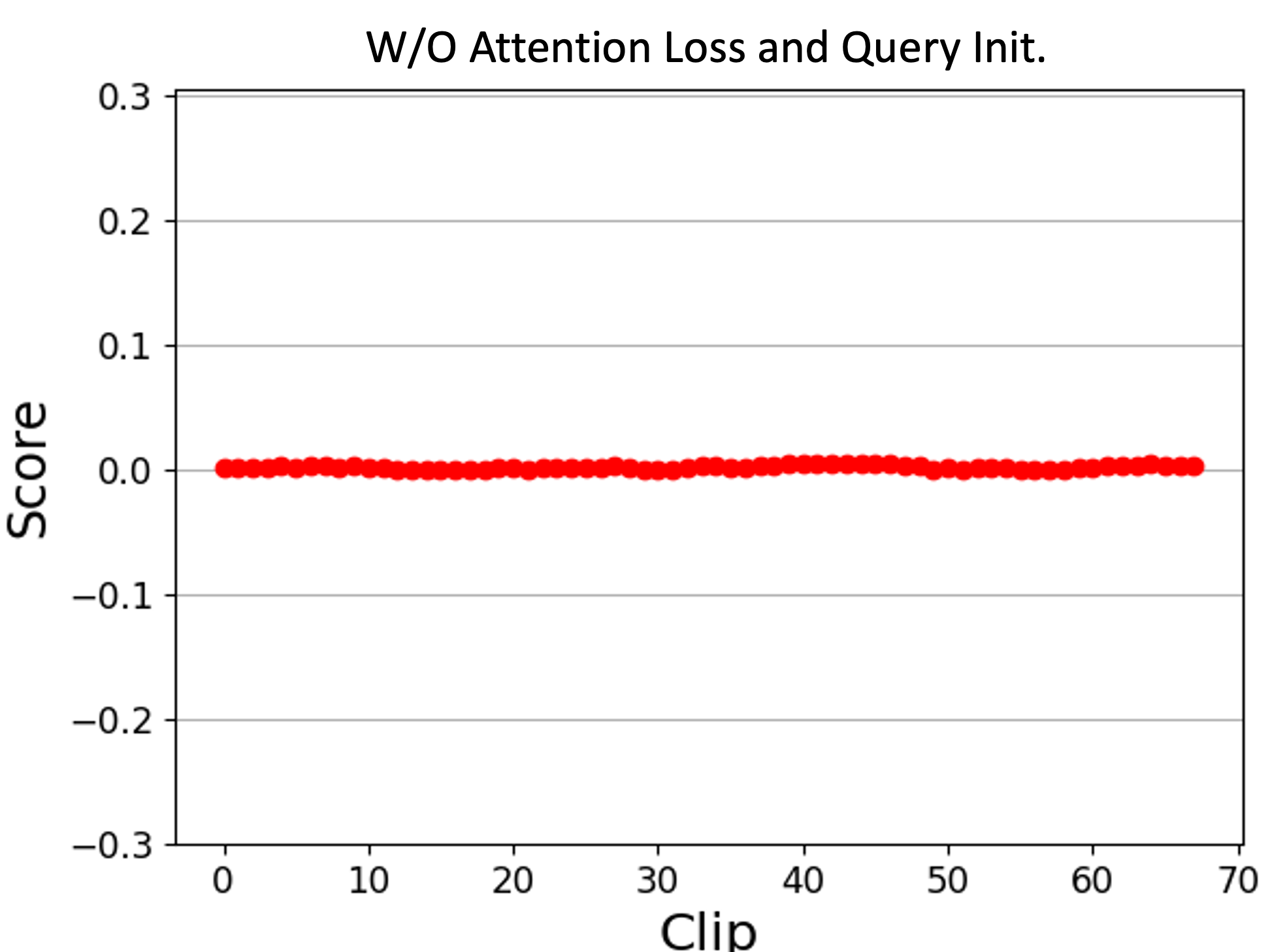}\
        \label{fig:4}
    }
        \subfigure[]{
        \includegraphics[width=0.23\linewidth,trim=0cm 0cm 0cm 0cm]{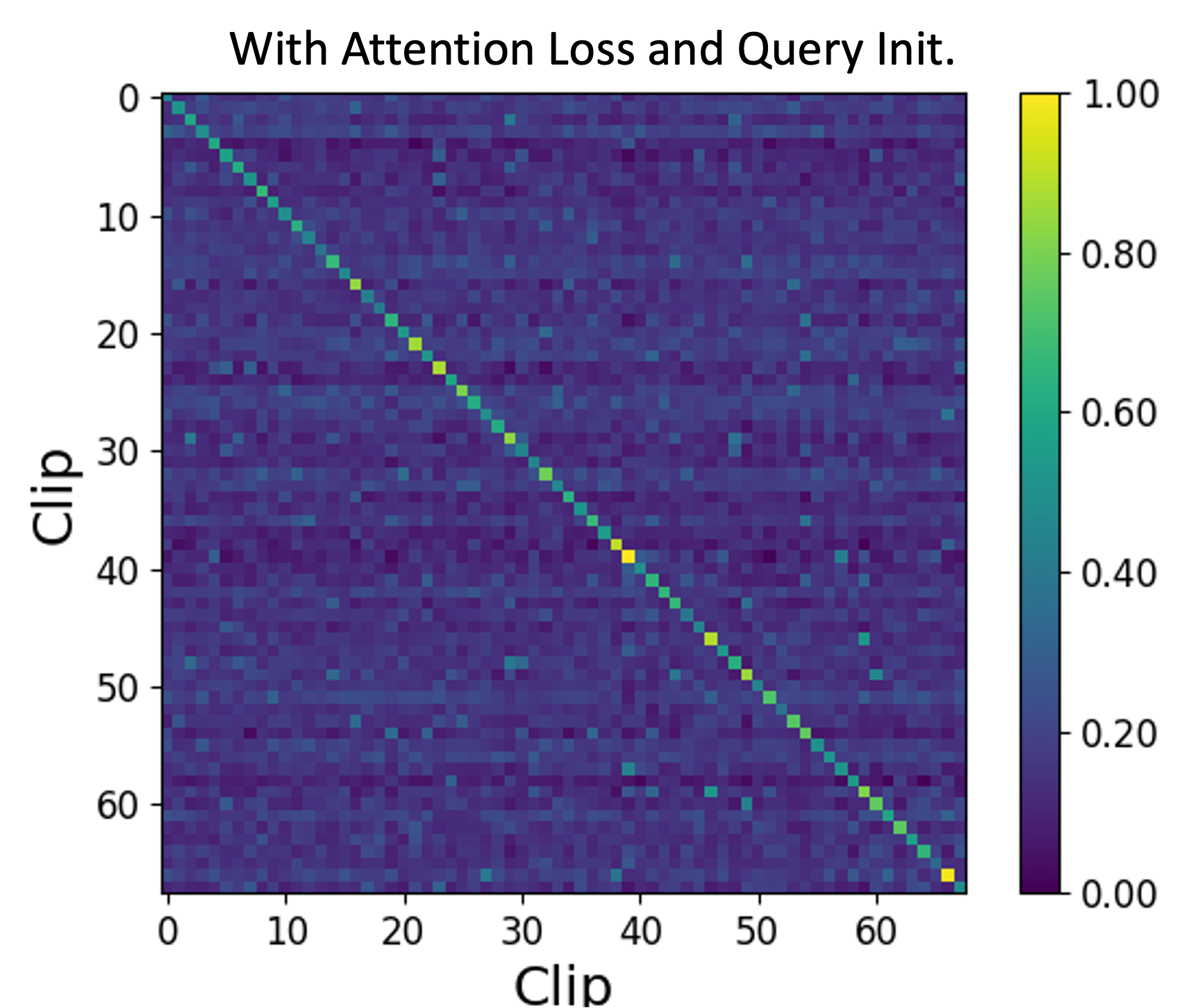}\
        \label{fig:1}
    }
        \subfigure[]{
        \includegraphics[width=0.23\linewidth,trim=1cm 0cm -1cm 0cm]{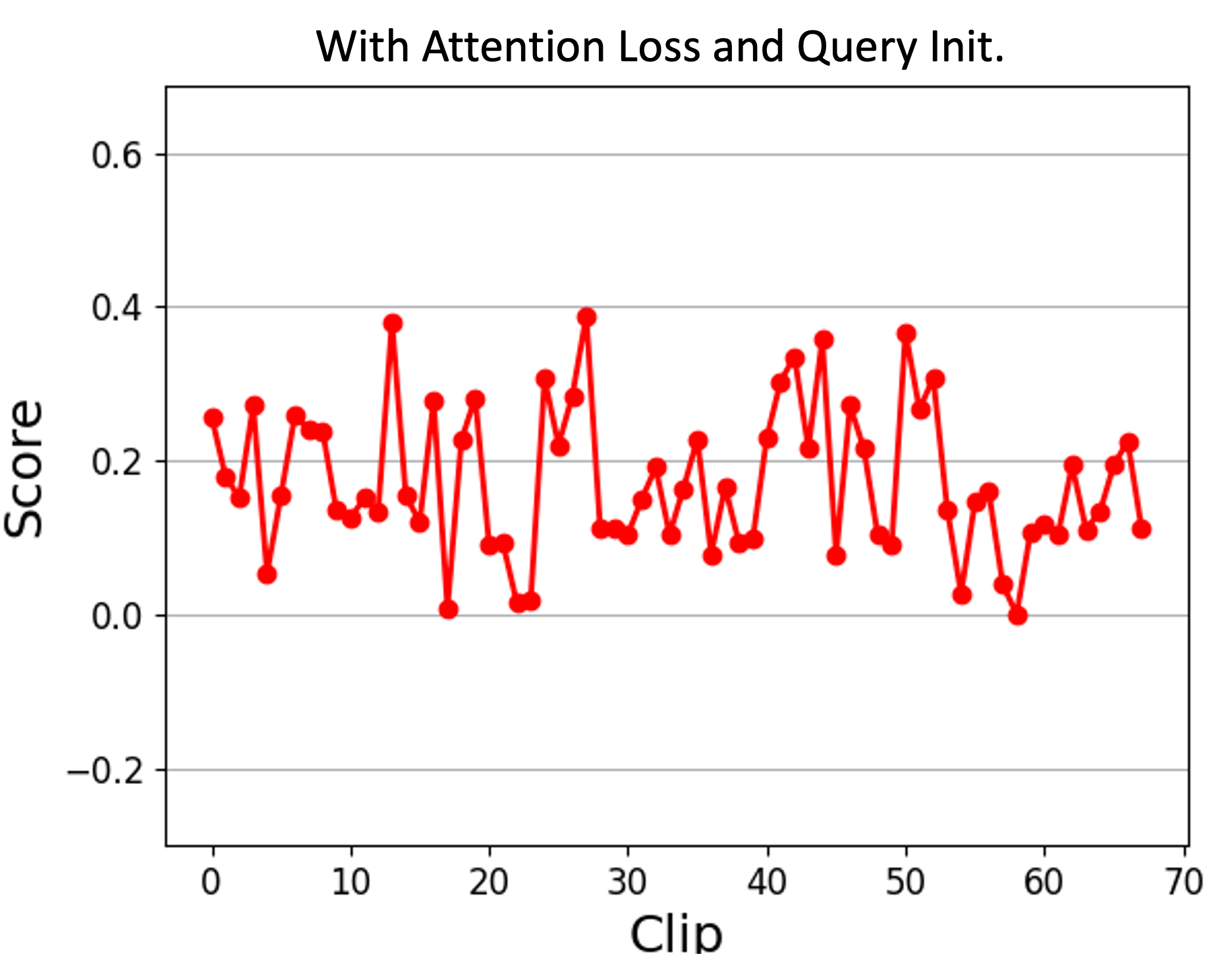}\
        \label{fig:2}
    }
    \caption{\textbf{\problemName problem of self-attention.} This figure shows the self-attention map \ref{fig:3} and \ref{fig:1} (ours) and visualization of segmented score of each clip \ref{fig:4} and \ref{fig:2} (ours). \ref{fig:3} and \ref{fig:4} represent the same action sequences, as do \ref{fig:1} and \ref{fig:2}. We can observe that in \ref{fig:3}, the self-attention map severely suffers from \problemName problem where \ref{fig:1} shows high correlations between queries.}
    \vspace{-10pt}
    \label{fig:a}
\end{figure}

Recently, approaches that use queries within an encoder/decoder transformer architecture network
\cite{carion2020endtoend,zhang2021temporal,bai2022action,du2023learning,xu2022likert} have been introduced in the AQA task due to their capabilities for long-term modelling, and because their decoder architecture is ideal for endowing the learnable queries with temporal semantic meanings. However, the model interpretability in long-term videos has not yet been satisfactory. One reason is that as each layer of the transformer is processed in long-term video, the decoder's self-attention may exhibit a \emph{skipping} phenomenon, as stated in \cite{kim2023self}. Temporal sequences lead the model to select shortcuts and skip decoder self-attention, thus preventing output degradation. The problem can be defined as \problemName in AQA, as shown in Figure \ref{fig:a}; We present our solution with the self-attention map in \ref{fig:1} and the segmented scores in \ref{fig:2}. In contrast, the opposite results are shown in \ref{fig:3} and \ref{fig:4}. Compared to \ref{fig:1}, which displays a distinct self-correlation diagonal of queries attention, Figure \ref{fig:3} exhibits a \problemName issue, resulting in smooth and averaged attention weights that deviate from the diagonal in the self-attention map. This results in a failure of interpretability in Figure \ref{fig:4} 's graph, where each clip has the same weight, whereas Figure \ref{fig:2} shows the opposite.


%

We propose introducing an Attention loss that facilitates mutual guidance between the self-attention and cross-attention maps to solve this issue. This is achieved by minimising the similarity between the two attention maps using KL divergence, ensuring that as the number of layers in the transformer decoder increases, the queries within the self-attention maintain a high correlation. Additionally, by altering the variance of the Gaussian distribution used to initialize the query embedding, the correlation of the self-attention map increases, as evidenced by a clearer diagonal in the self-attention map as in \ref{fig:1}. Encoding the position of the query and features is crucial to maintaining spatial and temporal encoding for interpretable sequences. Therefore, we propose to add positional encoding to the learnable queries to encode their temporal nature. We also explore positional encoding on the video features; however, we discover that because positional information has already been extracted in the backbone, providing only minimal additional temporal information. Moreover, complex positional encoding makes the features redundant, hindering network training and convergence.

Furthermore, inspired by how humans judge action quality and to enhance the rationality of the interpretability of the AQA model, we have developed a \regressionhead to substitute the conventional single-score regression approach. This module decouples the output of the DETR decoder into weight and score branches to align with the scoring logic of human judges in the real world. The final action score is obtained by calculating the weighted sum of the scores for each clip. Figure \ref{fig:inter} shows that our \regressionhead can parse clip-level scores and weights. In summary, our main contributions are as follows:
\begin{itemize}[itemsep=0pt]
  \item We propose a Query-based transformer decoder network for AQA, with positional query encoding to extract the clip-level feature with temporal semantic meanings. 
  \item We identify a \problemName issue in self-attention that causes interpretability failures and propose an Attention Loss and query initialization method to address it.
  \item To decouple the score into weight and score, We propose a split \regressionhead to improve the interpretability further.
  \item Our extensive experiments show that our network can extract interpretable features of temporal clips and achieve new state-of-the-art results on standard three long-term AQA benchmarks: Rhythmic Gymnastics (RG) \cite{zeng2020hybrid}, Figure Skating Video (Fis-V) \cite{xu2019learning} and LOng-form GrOup (LOGO) \cite{Zhang_2023_CVPR}.
\end{itemize}


\section{Related Work}
\paragraph{Action Quality Assessment}  Most existing research regards AQA as a regression task \cite{parmar2019action,pan2019action,xu2019learning,zhang2024auto,wang2021tsa}. \cite{pirsiavash2014assessing} first explored the AQA task by extracting spatiotemporal pose features of individuals and employing the L-SVR model to estimate and predict action scores. \cite{parmar2017learning} proposed three frameworks for evaluating the quality of Olympic event actions: C3D-SVR, C3D-LSTM, and LSTM-SVR. However, using only human pose information lacks modelling of external appearance information, such as splash size in diving. Additionally, LSTM lacks modeling of global relationships. Some other research focuses on addressing the uncertainty problem of AQA, \cite{tang2020uncertainty} proposed an Uncertainty Score Distribution Learning (USDL) method for enhancing action quality representation, which regards each action as an instance associated with a score distribution, thereby reducing the impact of inherent ambiguity in the score label. Another branch developed an AQA task as a ranking problem. \cite{Doughty_2019_CVPR} proposed a novel rank-aware loss function and trained it with a temporal attention module. \cite{yu2021group} proposed a group-aware regression tree (CoRe) method, which regresses the relative score and refers to other videos having similar attributes. However, these methods only regress a single score for the video sequences, which lack clip-level temporal semantic meanings. Regarding the interpretability of AQA, \citet{roditakis2021towards} utilized a self-supervised training technique and a differential cycle consistency loss to improve the temporal alignment and interpretability. \cite{farabi2022improving} stated that averagely aggregating the clip-level feature cannot capture the relative importance of clip-level features and proposed a weighted-averaging technique. Although these works focus on clip-level semantic meanings, they do not follow the scoring logic of human judges in the real world. Our work decouples clip-level features into weight and score, further enhancing the interpretability of AQA. 
\vspace{2pt}

\textbf{DETR in Video Understanding} DEtection TRansformer (DETR) was first introduced by \citet{carion2020endtoend}, which uses transformer architecture to capture complex relationships and dependencies in a set of data via learnable queries. Many works adapted DETR to video understanding tasks and demonstrated its ability in temporal modelling. \cite{zhang2021temporal} proposed a Temporal Query Network, which uses a query-response mechanism to regard each action clip in a video as a query. \cite{liu2022end} proposed a method that adopted Deformable DETR \cite{zhu2020deformable} for temporal action detection tasks, eliminating the proposal generation stage. \cite{kim2023self} first identified the temporal collapse problem in the temporal action detection task using a DETR-like structure and proposed a self-feedback method. Our work addresses a similar situation of temporal collapse but in the context of the AQA task and modifies the representation of the self-attention map and cross-attention map in the decoder.  \cite{bai2022action} proposed a Temporal Parsing Network (TPN) based on the DETR decoder for decomposing global features into temporal levels, which allows the network to parse temporal semantic meanings. However, TPN only evaluated short-term datasets, which lacked long-term video modelling capabilities. Our network is based on DETR and incorporates a query initialization module and attention loss on self-attention and cross-attention to prevent from \problemName.

\textbf{Long-term Video Understanding} Early work \cite{li2019beyond,gao2017tall} using RNNs for long-term video modelling. Recently, many works have adopted transformers due to their capabilities in long-term video modelling \cite{wang2021end,wu2022memvit}. In the AQA task, \cite{zeng2020hybrid} proposed a long-term video dataset and an ACTION-NET using GCN with a Context-aware attention module for temporal feature modelling. However, the performance of ACTION-NET is not satisfied on long-term datasets. \cite{xu2022likert} improved previous work by adopting the learnable query as a grade prototype and utilized a Likert Scoring Module for grade decoupling in the long-term video. Unlike other networks, our model adopts a transformer decoder structure, which can model long-term videos. Additionally, our attention loss and query initialization modules address the \problemName issue in modelling long-term videos, enhancing fine-grained feature extraction.
\vspace{-10pt}
\section{Method}
\begin{figure}[t]
  \centering
  \includegraphics[width=0.95\textwidth,height=0.47\textwidth]{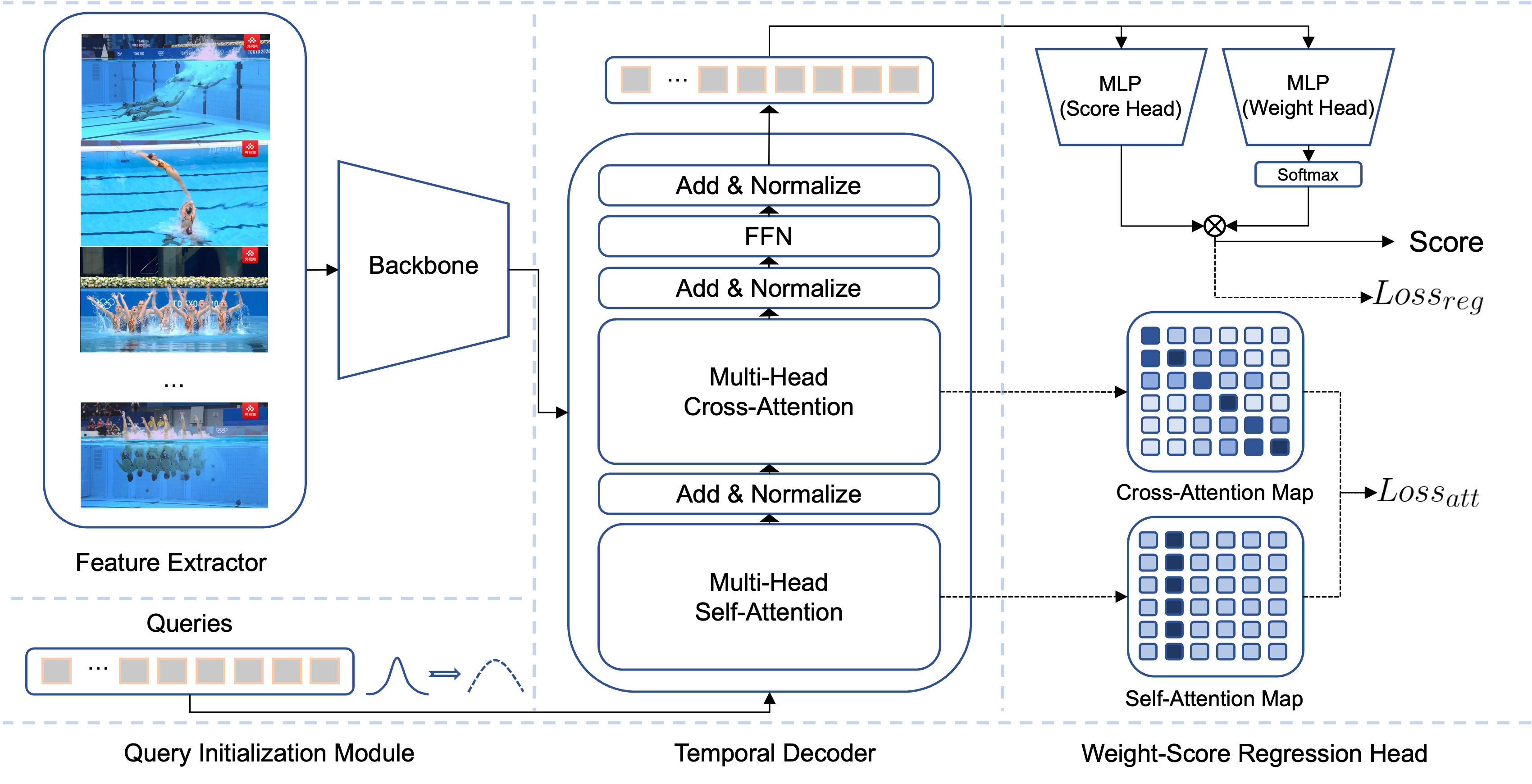}
  \caption{The overview architecture of our \networkName. The input video is divided into clips and fed into a backbone network. A temporal decoder models the clip-level features into temporal representations via learnable positionally encoded queries. The interpretable weight-score regression head can regress the final score by multiplying the weight and score of each clip. By minimizing the similarity between the self-attention map and cross-attention map, as well as query initialization, the problem of temporal collapse common in longer-term video sequences disappears and improves human interpretability.}
  \label{fig:network}
  \vspace{-10pt}
\end{figure}
Our network, shown in Figure \ref{fig:network}, consists of three modules: a \backbone used to extract the clip-level features of the input video, followed by a \temporaldecoder that encodes the attention relation between these features and a set of learnable positional encoded queries to extract temporal semantic features. Subsequently, the clip features from the temporal decoder are fed into a regression head module that decouples the weight and score of each action clip. The final score is obtained by multiplying the score of each clip by its weight and summing the results across all clips. The network is optimized using two loss functions: Attention Loss $Loss_{Att}$, the sum of the KL divergence between two attention maps at each layer, and an MSE Loss $Loss_{MSE}$ to measure the mean squared error. 

\textbf{Feature Extractor}
To extract the sequence or clip features from the input video $V$. we divide the video into $L$ non-overlapping clips, each containing $M$ consecutive frames, \(V=\{F^{i}\}_{i=1}^{L}\). We use two common feature extractors, Inflated 3D ConvNet (I3D) \cite{carreira2018quo} and Video Swin Transformer (VST) \cite{liu2021video,liu2021Swin} as our \backbone. The \backbone network is frozen, using the pre-extracted \backbone features as input to the \temporaldecoder and weight-score regression head network. The features obtained from $L$ clips are denoted as ${f^i}_{i=1}^L$, where each $f^i \in \mathbb{R}^d$.

\textbf{Temporal Decoder}
To encode the relationship of the temporal features, we use the decoder of a query-based positional transformer architecture \cite{carion2020endtoend}, as previously employing an encoder has been shown to reduce performance~\citet{bai2022action}. The decoder has layers with self-attention for processing query inputs and cross-attention, where queries interact with encoded clip features. The decoder has two layers, as having more layers leads to a deterioration of temporal skipping. A skip connection across the cross-attention layer is used to directly propagate the output of each self-attention layer to subsequent layers. The encoded clip features ${f^i}_{i=1}^L$ are taken from \backbone, and a set of learnable queries is used; the model is forced to set each query to correspond to a clip and assorted clip features. The cross-attention mechanism learns to match each action query with the memory. This matching process is achieved by computing attention scores between the query vector and all memory vectors, effectively selecting and focusing on the spatial and temporal information relevant to the query. However, different to the standard query initialization method in DETR, which takes 1 as variance, we adjusted the variance of the Gaussian distribution used to obtain embedding to increase the correlation between the queries in the self-attention map. Furthermore, in the vanilla DETR, sin or cos positional encodings for queries and memory are used to provide relational position information. However, in our decoder, only query positional encoding is used to eliminate the dependence on complex positional encodings or any prior knowledge, such as extracted temporal features from the backbone.

\textbf{Attention Loss} We observe that the model tends to skip the self-attention module after several training epochs. Consequently, the correlation between queries is lost, and each query carries a similar weight in self-attention. This phenomenon \problemName is illustrated in Figure \ref{fig:3}, where the self-attention map deviates from a desired diagonal line, which indicates a collapse of temporal information and a failure of temporal interpretability. Therefore, we propose the following \emph{Attention Loss} $Loss_{Att}$ to solve this. $A_S$ and $A_C$ are defined as the outputs of the self-attention and cross-attention layers, respectively. The self-attention map $L_S$ and $L_C$ are obtained by taking the matrix product of $A_S$, $A_C$ and its transpose. Finally, the Attention Loss $Loss_{Att}$ is formulated as equation \ref{eq:lossat}, where $D_{KL}$ is the Kullback-Leibler (KL) divergence, and N denotes the number of decoder layers in the \temporaldecoder. By using Attention Loss, each layer's self-attention and cross-attention in the transformer decoder maintain a high correlation with the previous layer until the output of the final layer, thereby mitigating the problem of temporal smoothing and resolving issues of interpretability failure.

\begin{minipage}{.48\linewidth}
\begin{equation}
L_S=softmax({A_SA_S}^\intercal)
\end{equation}\end{minipage}%
\begin{minipage}{.48\linewidth}

\begin{equation}
L_C=softmax({A_CA_C}^\intercal)
\end{equation}
\end{minipage}

\begin{equation}
Loss_{att}=\sum_{n=1}^{N}D_{KL}(L_S^n||L_C^n)
\label{eq:lossat}
\end{equation}

\textbf{Weight-Score Regression Head} To mimic human-like scoring in our AQA model. We propose a \regressionhead to decouple each action clip's weight and quality. This is achieved by using parallel weight and score heads, each implemented as an MLP layer, with the weight head employing a softmax function to output values between 0 and 1. The final $score$ is the sum of each clip's score multiplied by its $weight$, shown in equation \ref{eq:weight_score}, where $K$ is the number of clips. 

\textbf{Model Training}
We follow the prior work of AQA \cite{yu2021group,wang2021tsa,Zhang_2023_CVPR,xu2022likert} alongside our Attention Loss to utilize  Mean Square Error Loss (MSE) to minimise the predicted value and ground truth value as shown in equation \ref{eq:reg}, where $y$ is the predicted value and $\hat{y}$ is the ground truth value. The final loss can then be described as in equation \ref{eq:overall} where $\lambda_{reg}$ and $\lambda_{Att}$ are the weights for MSE loss and attention loss. 

\begin{minipage}{.48\linewidth}
\begin{equation}
    score=\sum_{k=1}^{K}weight_k\cdot score_k
    \label{eq:weight_score}
\end{equation}
\end{minipage}%
\begin{minipage}{.48\linewidth}

\begin{equation}
Loss_{reg} = \frac{1}{n} \sum_{i=1}^{n} (y_i - \hat{y}_i)^2
\label{eq:reg}
\end{equation}
\end{minipage}

\begin{equation}
Loss_{all}=\lambda_{reg}Loss_{reg}+\lambda_{att}Loss_{att}
\label{eq:overall}
\end{equation}

\vspace{-10pt}
\section{Experiment}
\subsection{Datasets and Metrics}
Experiments are conducted on three widely used long-term AQA benchmarks, including Rhythmic Gymnastics \cite{zeng2020hybrid}, LOng-form GrOup \cite{Zhang_2023_CVPR} and Figure Skating Video \cite{xu2019learning} to evaluate our model. To be consistent with prior research \cite{Doughty_2019_CVPR,yu2021group,roditakis2021towards,farabi2022improving,zhang2021temporal}, we adopted two metrics in our experiment, the \textbf{Spearman's rank correlation (SRCC)} and \textbf{Relative L2 distance (R-$\ell$2)}. Spearman's rank correlation measures the correlation between two sequences containing ordinal or numerical data, with values ranging from -1 to 1 (higher values indicating stronger correlation) as shown in \ref{eg:srcc}. In contrast, Relative L2 distance calculates the Euclidean distance between corresponding elements of two sequences, providing a measure of their dissimilarity (lower values indicating better similarity) as shown in \ref{eg:rl2}.
\begin{minipage}{.47\linewidth}
\centering
\begin{equation}
\rho =\frac{\sum_{i}^{}(p_i-\bar{p})(q_i-\bar{q})}{\sqrt{\sum_{i}(p_i-\bar{p})^2\sum_{i}(q_i-\bar{q})^2}}
\label{eg:srcc}
\end{equation}\end{minipage}%
\begin{minipage}{.47\linewidth}
\centering
\begin{equation}
R\text{-}\ell2=\frac{1}{N}\sum_{N}^{n=1}(\frac{\left|y_n-\hat{y}_n\right|}{y_{max}-y_{min}})^2
\label{eg:rl2}
\end{equation}
\end{minipage}

\textbf{Rhythmic Gymnastics (RG). \cite{zeng2020hybrid}} 
The RG dataset contains video sequences of four distinct types of gymnastics routines: ball, clubs, hoop, and ribbon. Each action class comprises 200 training samples and 50 evaluation samples, with each sample approximately 1 minute and 35 seconds in duration. Each class is trained as a model, adhering to the practices described in \cite{zeng2020hybrid, xu2022likert}.

\textbf{Figure Skating Video (Fis-V). \cite{xu2019learning}} The Fis-V dataset contains 500 videos of figure skating videos with an average length of 2 minutes and 50 seconds. We followed the previous work, which had 400 videos for training and 100 videos for testing. Fis-V provides two labels: Total Element Scores (TES) and Total Program Component Score (PCS). Two individual models are trained to predict scores for two classes.

\textbf{LOng-form GrOup (LOGO). \cite{Zhang_2023_CVPR}} The LOGO dataset is a multi-person long-term video dataset with 150 samples for training and 50 for testing. Each video sequence is approximately 3 and a half minutes in length. To our knowledge, LOGO has the longest video lengths among existing AQA datasets.

\subsection{Implementation Details}
We adopt Inflated 3D ConvNet (I3D) \cite{carreira2018quo}, and Video Swin Transformer (VST) \cite{liu2021video,liu2021Swin} pretrained on Kinetics as our video feature extractor. Note that we only utilize the \backbone to extract features and do not train it. The clip and query sizes for RG, FIS-V, and LOGO are set to 68, 136, and 48, respectively. These parameters are determined based on the length of the videos and extensive comparison experiments, demonstrating optimal performance. The Adam optimizer is adopted with a learning rate $1 \times 10^{-4}$ and a batch size of 48. The output dimension of the transformer decoder is 1024, with each decoder having 4 heads and 2 layers, each layer applying a dropout rate of 0.7. For the weight-score regression head, each module consists of three MLP layers and a softmax layer after the weight branch.

\subsection{Results and Analysis}

The \networkName is compared with state-of-the-art methods on three benchmarks. Results on the RG and Fis-V dataset are shown in Table \ref{tab:RG_results}; our model outperforms the current state-of-the-art method GDLT \cite{xu2022likert} on RG, which uses the same image features on the four subclasses, achieving an average improvement of 0.077. Furthermore, our model also outperforms the current state-of-the-art method GDLT \cite{xu2022likert} by an average of 0.027 on Fis-V dataset of two subclasses, TES and PCS. On the LOGO dataset, our model achieves state-of-the-art as shown in Table \ref{tab:LOGO_results} and outperforms prior methods \cite{Zhang_2023_CVPR} by 0.220 using VST as the \backbone and 0.099 using I3D as the \backbone on SRCC. 
\begin{table}[ht]

  \begin{center}
  \renewcommand{\arraystretch}{1.1}
\scriptsize
\begin{tabular}{c|c|ccccc|ccc}
\hline
\multirow{2}{*}{Methods}                       & \multirow{2}{*}{Feature Extractor}                                                           & \multicolumn{5}{c|}{RG (SRCC$\uparrow$)}                                                                      & \multicolumn{3}{c}{Fis-V (SRCC$\uparrow$)}                                                         \\ \cline{3-10} 
                                               &                                                                                     & Ball             & Clubs            & Hoop             & Ribbon           & Avg.             & TES                       & PCS                       & Avg.                      \\ \hline\hline
SVR \cite{parmar2017learning} & C3D \cite{tran2015learning}                                       & 0.357            & 0.551            & 0.495            & 0.516            & 0.483          & \multicolumn{1}{l}{0.400} & \multicolumn{1}{l}{0.590} & \multicolumn{1}{l}{0.501} \\ \cline{1-2}
MS-LSTM                                        & I3D \cite{carreira2018quo}                                         & 0.515            & 0.621            & 0.540            & 0.522            & 0.551          & -                         & -                         & -                         \\ \cline{2-2}
\cite{xu2019learning}         & VST \cite{liu2021video}                                            & 0.621            & 0.661            & 0.670            & 0.695            & 0.663          & 0.660                     & 0.809                     & 0.744                     \\ \cline{1-2}
ACTION-NET                                     & I3D\cite{carreira2018quo}+ResNet\cite{he2016deep} & 0.528            & 0.652            & 0.708            & 0.578            & 0.623          & -                         & -                         & -                         \\ \cline{2-2}
\cite{zeng2020hybrid}         & VST\cite{liu2021video}+ResNet\cite{he2016deep}    & 0.684            & 0.737            & 0.733            & 0.754            & 0.728          & 0.694                     & 0.809                     & 0.757                     \\ \cline{1-2}
GDLT \cite{xu2022likert}      & VST \cite{liu2021video}                                            & 0.746            & 0.802            & 0.765            & 0.741            & 0.765          & 0.685                     & 0.820                     & 0.761                     \\ \hline
\textbf{Ours}                                  & VST \cite{liu2021video}                                            & \textbf{0.823} & \textbf{0.852} & \textbf{0.837} & \textbf{0.857} & \textbf{0.842} & \textbf{0.717}            & \textbf{0.858}            & \textbf{0.788}            \\ \hline
\end{tabular}
  \end{center}
  \caption{Spearman's rank correlation coefficient performance comparison on \textbf{Rhythmic Gymnastics (RG)} and \textbf{Figure Skating Video (Fis-V)} dataset. Avg. is the average SRCC for all subclasses. The higher SRCC suggests better performance.}
  \label{tab:RG_results}
  \vspace{-6pt}
\end{table}

\begin{table}[ht]
  \begin{center}
    \renewcommand{\arraystretch}{1.1}
      \begin{tabular}{c|cc|cc}
        \hline
        \multirow{2}{*}{Methods} & \multicolumn{2}{c|}{I3D \cite{carreira2018quo}} & \multicolumn{2}{c}{VST \cite{liu2021video}}          \\ \cline{2-5} 
                                 & SRCC $\uparrow$       & R-$\ell$2(×100) ↓  & SRCC $\uparrow$              & R-$\ell$2(×100) ↓    \\ \hline\hline
        USDL \cite{tang2020uncertainty}                    & 0.426   & 5.736        & 0.473          & 5.076          \\ \cline{1-1}
        CoRe \cite{yu2021group}                    & 0.471   & 5.402        & 0.500          & 5.960          \\ \cline{1-1}
        TSA \cite{xu2022finediving}                  & 0.452   & 5.533        & 0.475          & 4.778          \\ \cline{1-1}
        ACTION-NET \cite{zeng2020hybrid}            & 0.306   & 5.858        & 0.410          & 5.569          \\ \cline{1-1}
        USDL-GOAT \cite{Zhang_2023_CVPR}              & 0.462   & 4.874        & 0.535          & 5.022          \\ \cline{1-1}
        TSA-GOAT \cite{Zhang_2023_CVPR}                & 0.486   & 5.394        & 0.484          & 5.409          \\ \cline{1-1}
        CoRe-GOAT \cite{Zhang_2023_CVPR}               & 0.494   & 5.072        & 0.560          & 4.763          \\ \hline
        \textbf{Ours} & \textbf{0.593}       & \textbf{1.220}             & \textbf{0.780} & \textbf{1.745} \\ \hline
        \end{tabular}
    \end{center}
    \caption{Performance comparison on \textbf{LOGO} dataset. The higher the SRCC, the lower R-$\ell$2, it suggests better performance.}
      \label{tab:LOGO_results}
      \vspace{-10pt}
\end{table}

\paragraph{Ablation Study}
Experiments were conducted on three settings to compare the effects of attention loss, query positional encoding, and query initialization. As shown in table \ref{tab:ablation}, only use data-decoder without any other proposed modules. Our proposal has only 0.628 SRCC as the baseline. Adding the attention loss significantly improved the performance by 28.5\%, which verifies that our proposed attention loss improves the interpretability and the SRCC results. With the addition of only query positional encoding, the results improved to 0.810. Finally, our query initialization module further enhanced the performance by around 4\%, showing that using high variance allows query vectors to have a wider diversity and dispersion of initialization states and improve the overall performance.

\begin{table}[h]
  
  \begin{center}
  \renewcommand{\arraystretch}{1}
    \begin{tabular}{c|ccc|c}
    \hline 
    Module  & Attention Loss & Query PE & Query Init. & SRCC $\uparrow$ \\ \hline\hline
    Baseline & $\times$                  & $\times$         & $\times$              &  0.628  \\
     & $\checkmark$              & $\times$         & $\times$              &  0.807   \\
             & $\checkmark $             & $\checkmark$        & $\times$           &  0.810 \\ \hline
    \textbf{Ours}     & $\checkmark$              & $\checkmark$        & $\checkmark$                    &  \textbf{0.842}\\ \hline  
    \end{tabular}
  \end{center}
  \caption{\textbf{Ablation study} on the average performance of four labels in the Rhythmic Gymnastics (RG) dataset across various modules.}

  \label{tab:ablation}
  \vspace{-10pt}
\end{table}
\begin{table}[htbp]
\begin{minipage}[htbp]{.45\linewidth}
  \begin{center}
  \renewcommand{\arraystretch}{1}
    \begin{tabular}{c|cc|c}
    \hline
    Methods & Query & Memory & SRCC  \\ \hline\hline
    Baseline            &  $\times$     &   $\times$     & 0.758 \\
                        &  $\times$    &  \checkmark      & 0.778   \\
                        &  \checkmark     &  \checkmark      & 0.751 \\ \hline
    Ours                & \checkmark      &  $\times$      & \textbf{0.824} \\ \hline
    \end{tabular}
    \end{center}
    \caption{Effect of Positional Encoding on RG dataset, where SRCC results take the average of the four labels.}
    \label{tab:pe}
    \end{minipage}
\hfill
\begin{minipage}[htbp]{.45\linewidth}
  \begin{center}
  \renewcommand{\arraystretch}{1}

\begin{tabular}{c|c}
\hline
Variance Init. & SRCC  \\ \hline\hline

0.5      & 0.810 \\
1        &  0.810     \\ 
3        &  0.811     \\ 
5       &  \textbf{0.820}    \\ \hline
\end{tabular}

\end{center}
\caption{Effect of Query Variance Initialization on RG dataset, where SRCC results take the average of the four labels}
\label{tab:queryinit}
\end{minipage}
\end{table}
\vspace{-12pt}

\paragraph{Effect of position encoding}
Different positional encoding methods in the \temporaldecoder are compared in Table \ref{tab:pe}. We find that only adopting query positional encoding outperformed other methods, while incorporating query and memory positional encoding harm the final performance. This phenomenon is probably because, in the AQA task, we primarily focus on modelling learnable queries using the DETR decoder and endowing these queries with temporal semantic meanings through the decoder structure. However, the temporal information of the memory is already extracted by the \backbone, and using only the query positional encoding avoids unnecessary computation and potential information redundancy. This approach ensures that these queries can effectively capture and express the key action quality indicators in the video, thereby enhancing the scoring performance and the model's interpretability.
\vspace{-8pt}
\paragraph{Effect of variance in query initialization module}

In the \temporaldecoder, the impact of different variances in query initialization on the \problemName problem and final SRCC results is compared as shown in Table \ref{tab:queryinit} specifically, using a larger variance to initialize the query embedding results in a more compact diagonal pattern in the self-attention map, which represents higher correlation between action queries. Furthermore, initializing variance boosts the final SRCC results, as shown in Table \ref{tab:queryinit}. Different initialization variance values are compared in the experiment, and it is concluded that the SRCC is highest when using a larger variance.

\begin{figure}[h]
  \centering
  \includegraphics[width=0.98\textwidth,height=0.38\textwidth,trim=0cm 7cm 0cm 0cm]{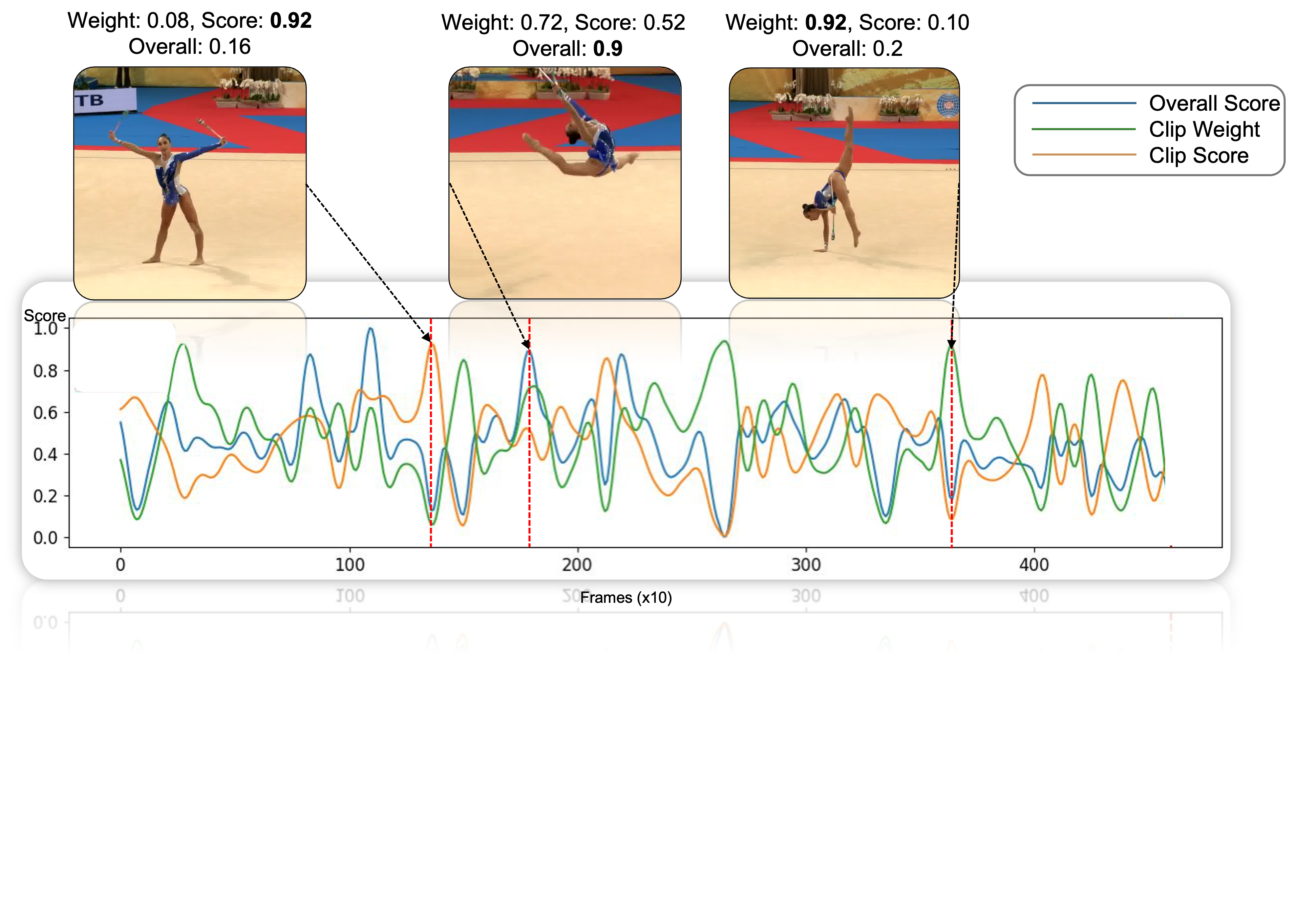}
  \caption{Visualization of our clip-level weight-score regression method on RG dataset.}
  
  \label{fig:inter2}
  \vspace{-10pt}
\end{figure}
\vspace{-8pt}
\paragraph{Sequence Interpretability}
To replace the single-score regression method and follow the scoring logic of human judges, we decoupled each clip's score into weight and score. Figure \ref{fig:inter} (a) shows that in the first clip, despite high completion quality and synchronization, the weight (difficulty) is low. In the second clip, the action has a high weight and score. However, the movement quality is low in the last clip, although the difficulty of the action is high (high weight). Figure \ref{fig:inter2} shows another visualization of the clip-level weight-score regression method on the RG dataset, where the green line represents the weight (difficulty) of the current frame, the yellow line represents the score (quality) of the current frame, and the blue line represents the combined score. Empirical evidence demonstrates that our weight-score regression module is effective in enhancing the interpretability of AQA. 

\vspace{-8pt}
\subsection{Conclusion}

In this work, we propose a novel framework to enhance interpretability in long-term AQA tasks, addressing the \problemName issue which fails interpretability. A novel attention loss function and a query initialization module are proposed, and the impact of different positional encodings is explored. Our approach also includes a weight-score regression module that decouples each clip's action score into weight and score, facilitating a fine-grained and interpretable assessment that makes AQA scoring more meaningful and informative. Demonstrating effectiveness, our model achieves state-of-the-art results on three AQA benchmarks and effectively parses clip-level semantic meanings through interpretability results. In our future work, we will further explore interpretability evaluation methods from both qualitative and quantitative perspectives, aiming to enhance the evaluation process in interpretable AQA tasks by making it more comprehensive, and accessible for analysis.

\bibliography{egbib}
\end{document}